\documentclass[letterpaper, 10 pt, conference]{ieeeconf}
\IEEEoverridecommandlockouts          
\usepackage{algorithm}
\usepackage{algorithmicx}
\usepackage[noend]{algpseudocode}
\usepackage{amsmath, amsfonts, bm, amssymb, tabularx}
\usepackage{latexsym, mathtools}
\usepackage{amsthm}
\usepackage[caption=false]{subfig}
\usepackage{ifthen}
\usepackage{hyperref}\hypersetup{colorlinks=true, unicode=true, linkcolor=[rgb]{0.10,0.05,0.67}, citecolor=[rgb]{0.10,0.05,0.67}, filecolor=[rgb]{0.10,0.05,0.67}, urlcolor=[rgb]{0.10,0.05,0.67}}
\usepackage{algorithm}
\usepackage{Shorthands}
\usepackage{cleveref}
\usepackage{tablefootnote}
\usepackage{caption}
\usepackage[numbers, sort&compress]{natbib}
\title{\LARGE \bf
Uncertainty-Aware Deployment of Pre-trained Language-Conditioned Imitation Learning Policies
}

\author{Bo Wu$^{1}$, Bruce D. Lee$^1$, Kostas Daniilidis$^1$, Bernadette Bucher$^{1,2}$, Nikolai Matni$^1$
\thanks{$^{1}$GRASP Lab, University of Pennsylvania:
        {\tt\small \{bobwu, brucele, kostas, nmatni\}@seas.upenn.edu}}%
\thanks{$^{2}$Robotics Department, University of Michigan: {\tt\small bucherb@umich.edu}}
\thanks{BL and NM gratefully acknowledge support from NSF Award SLES 2331880, NSF CAREER award ECCS 2045834, and NSF EECS 2231349.}
}

\begin{document}

\maketitle
\thispagestyle{empty}
\pagestyle{empty}

\begin{abstract}

Large-scale robotic policies trained on data from diverse tasks and robotic platforms hold great promise for enabling general-purpose robots; however, reliable generalization to new environment conditions remains a major challenge. 
Toward addressing this challenge, we propose a novel approach for uncertainty-aware deployment of pre-trained language-conditioned imitation learning agents. Specifically, we use temperature scaling to calibrate these models and exploit the calibrated model to make uncertainty-aware decisions by aggregating the local information of candidate actions.
We implement our approach in simulation using three such pre-trained models, and showcase its potential to significantly enhance task completion rates. The accompanying code is accessible at the link: \href{https://github.com/BobWu1998/uncertainty_quant_all.git}{\text{https://github.com/BobWu1998/uncertainty\_quant\_all.git}}

\end{abstract}




\section{Introduction}
\label{sec:introduction}

Inspired by advancements in general-purpose natural language processing and computer vision models, the robotics community has invested considerable effort in developing generalist decision making agents that operate across various robotic platforms and perform a broad spectrum of tasks \citep{yang2023foundation}. These so-called ``foundation models'' are typically trained once on a large and diverse dataset, then deployed in a specific setting with minimal task-specific fine-tuning. Such pre-trained models have been shown to generalize in many ways such as to new tasks,  new robot platforms, and new  environments~\citep{dasari2019robonet,ebert2021bridge,brohan2022rt,brohan2023rt}.

However, the extent of these agents' generalization abilities remains poorly understood. A significant barrier hindering the generalization abilities and our understanding of them is the lack of well-calibrated, task-specific uncertainty estimates. This gap precludes the possibility of making uncertainty-aware decisions. Our work aims to close this gap by proposing a calibration enabled uncertainty-aware decision making protocol to boost the success rate for language-conditioned imitation learning (IL) policies. 
\vspace{-2pt}
\subsection{Contributions}

\begin{figure}[t]
    \centering
    \includegraphics[width=1\linewidth]{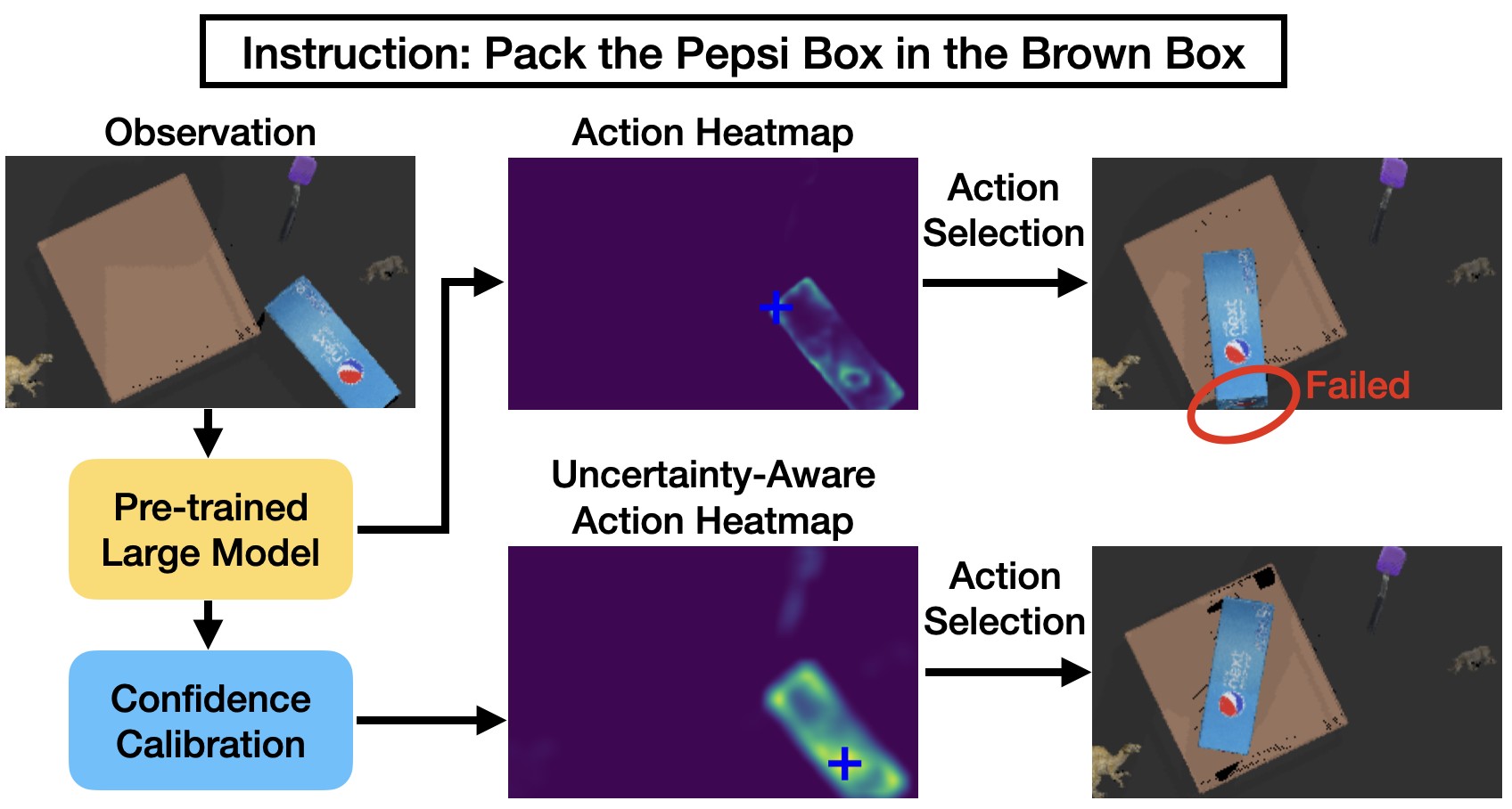}
    \caption{Overview of uncertainty-aware action selection on top of pre-trained models. The blue cross indicates the maximum score in the heatmap. The upper path illustrate the standard deployment of pre-trained IL policies at test time in which the output of the pre-trained model directly predicts the action. The bottom path indicates our proposed approach, in which the model is calibrated using a small dataset of expert demonstrations from the task of interest before selecting the action in an uncertainty-aware manner.}
    \vspace{-16pt}
    \label{fig: title figure}
\end{figure}

We propose a straightforward yet effective modification to enhance the generalization capabilities of pre-trained language-conditioned imitation learning agents for robotic manipulation. The modification is composed of two components, outlined in \Cref{fig: title figure}:

\begin{itemize}
    \item A calibration\footnote{We refer to calibration in the statistical sense \citep{guo2017calibration}.} step, tailored to the task of interest, that refines the model's outputs to generate confidence scores that approximate the correctness likelihood for imitating the expert. 
    \item An uncertainty-aware action selection technique using the probability distribution output by the calibrated model (See \Cref{alg: uncertainty aware action selection v2}).
\end{itemize}
Each of these steps is novel, and may be of independent interest. However, we view our 
primary contribution as the holistic integration of these two components into the decision making pipeline for language-conditioned IL policies. 

We instantiate our approach on several robotic manipulation models, including the Perceiver-Actor model \citep{shridhar2023perceiver}, the RVT model \citep{goyal2023rvt}, and the CLIPort model \citep{shridhar2022cliport} \emph{without additional training}, and show that using a small calibration dataset of expert demonstrations from the target task can lead to a substantial improvement in task completion rate. Through comprehensive experimentation, we identify scenarios where integrating uncertainty quantification into the decision making process is most beneficial. Our experiments also find that existing pre-trained models provide poorly calibrated likelihood estimates when instantiated on any particular task of interest. Therefore, appropriately calibrating the model for the task of interest is essential to reap the rewards of uncertainty-aware decision making. 
 
\subsection{Related Work}

\paragraph{Generalist Robotic Policies} 
Early efforts \citep{dasari2019robonet} introduce a diverse and extensive robotics dataset, and use this dataset to train visual foresight and supervised inverse models. By conditioning IL agents with a description of the task to complete, subsequent work uses such diverse datasets to enable sample-efficient IL \citep{ebert2021bridge, ding2019goal, stepputtis2020language, lynch2020language}. Recent developments have embraced transformer-based architecture to train multi-modal decision making agents
\citep{reed2022generalist, driess2023palm} and language-conditioned IL policies for robotic manipulation 
\citep{brohan2022rt, brohan2023rt, shridhar2022cliport, shridhar2023perceiver, goyal2023rvt}. Efforts have also been made to extend to more general prompt conditioning using images and videos \citep{jiang2022vima}. In \citet{shah2023vint}, the authors explore the development of a generalist visual navigation model. For a review of generalist decision making agents, see \citet{yang2023foundation}. Our work improves the performance of these generalist robotic policies by incorporating calibrated uncertainty estimates to enable more informed decisions at inference time.

\paragraph{Uncertainty Quantification} 
For robotic decision making, confidence scores quantifying the model output uncertainty enable downstream formulations of safe decisions in robotic planning problems \citep{lindemann2023safe, dixit2023adaptive, lekeufack2023conformal}. However, it has been shown that the confidence scores output by modern classification models are poorly calibrated in that they are not representative of the true correctness likelihood \citep{guo2017calibration}. 
While numerous approaches to improving calibration post-hoc exist, such as temperature scaling \citep{guo2017calibration} and conformal prediction \citep{shafer2008tutorial}, these techniques are seldom applied in the realm of pre-trained language-conditioned imitation learning policies to enhance generalization performance. A notable exception is the recent work by \citet{ren2023robots} which applies conformal prediction to language-based planners to request operator clarification in the face of ambiguity. In contrast, we consider a setting in which there is no ambiguity in the task-conditioning, but rather the pre-trained model is imperfect. We calibrate the uncertainty of this pre-trained model, and use it to select actions in an uncertainty-aware manner. For further discussion of uncertainty quantification of foundation models for robotic decision making, see \citet{firoozi2023foundation}.

\section{Approach}
\label{sec: method overview}
\subsection{Background: Language-conditioned IL for Robotic Manipulation}
\label{sec: il for robotic manipulation}

We consider robotic manipulation problems in a discrete-time decision making context. Let $o_t \in \calO$ be the observation of the world at time $t$ and
 $a_t \in \calA$ be an action applied to the robot at time $t$. Here, $\calA$ is  a finite, discretized action space. 
 
Language-conditioned IL assumes access to $n$ expert demonstrations $\curly{\zeta_1, \dots, \zeta_n}$ accompanied by language-conditioning $\curly{c_1, \dots, c_n}$. Each demonstration consists of a trajectory of observations, and the corresponding actions taken by an expert. The language-conditioning $c_i$  provides information about the task which the corresponding expert demonstration is completing. Using these demonstrations, the learner searches for a policy which imitates the expert behavior for the given language conditioning. 

As the action space is discrete, the policy operates as a classifier for the observation-language pair. In particular, given some observation $o\in\calO$ and language-conditioning $c$, the policy selects an action as 
\begin{align}
    \label{eq: standard policy}
    \hat \pi(o, c) = \argmax_{a \in \calA} \hat f_{j(a)}(o, c).
\end{align}
 Here, $j: \calA \to [\abs{\calA}]$ is a function enumerating $\calA$,  and $\hat f$ is a function which takes in an observation $o$ and a language conditioning $c$ to output a vector of logits $\hat f(o, c)$  with $\hat f_{j(a)}(o, c)$ representing its $j(a)^{\mathrm{th}}$ element. 
 
 A policy of the form mentioned above may be determined from the expert demonstrations and language-conditioning using behavior cloning. In particular, the learner may choose $\hat f$ belonging to some function class to minimize the cross entropy loss, as in \citet{brohan2022rt} and \citet{shridhar2023perceiver}. The resultant function,  $\hat f$, serves as a pre-trained model, which can be deployed on numerous different tasks. 
In particular, given some new language conditioning $c_{n+1}$, it is deployed to select the robot's actions as $a_t = \hat \pi(o_t, c_{n+1}).$ In general, the target task need not be one of the tasks used for training; we may have $c_{n+1} \notin \curly{c_1, \dots, c_n}$. However, many of our experimental results are focused on the setting where $c_{n+1} \in \curly{c_1, \dots, c_n}$. 

Pre-trained models using the aforementioned strategy to select an action neglect pertinent information output by the model about the distribution of potential actions. This is in part because such pre-trained models tend to be poorly calibrated and fail to reflect the actual correctness likelihood of prediction matching the expert demonstration. Accordingly, we address these problems by calibrating the model and then following an uncertainty-aware action selection strategy, which is discussed in Section \ref{sec: calibration} and Section \ref{sec: uaas}.

\subsection{Target Task Calibration}
\label{sec: calibration}

The logits can be viewed as a representation of a probability distribution $p$ over possible actions by applying the softmax function $\sigma_{SM}$ to the output of the model $\hat f$: 
$p = \sigma_{SM}(\hat f(o,c)).$ While $p$ is a viable probability distribution, it may not be well-aligned with the true correctness likelihood for imitating the expert, especially for the specific task $c_{n+1}$.

To improve the alignment between the distribution $p$ and the correctness likelihood for task $c_{n+1}$, we collect a calibration set from task $c_{n+1}$. In particular, we assume that the target language-conditioning $c_{n+1}$ is drawn randomly from some distribution $c_{n+1} \sim \calD_{\mathsf{test}}$, and that we have access to a calibration set of expert demonstrations $\curly{\zeta_{1}^{\mathsf{cal}}, \dots, \zeta_{k}^{\mathsf{cal}}}$ along with their associated conditioning $\curly{c_{1}^{\mathsf{cal}}, \dots, c_{k}^{\mathsf{cal}}}$, where $c_{i}^{\mathsf{cal}}$ are drawn from the same distribution as $c_{n+1}$: $c_{i}^{\mathsf{cal}} \sim \calD_{\mathsf{test}}$. Calibrated models may be obtained from this set with various methods, but we propose using temperature scaling \cite{guo2017calibration}. 
Specifically, we determine the optimal temperature parameter $\hat T$ as the value of $T$ which minimizes the cross entropy loss of the model $\frac{1}{T} \hat f$ over the calibration set.
Given this temperature parameter, we define the calibrated model outputting the probability scores as $\tilde p = \tilde f(o, c) = \sigma_{SM}\left(\frac{\hat f(o, c)}{\hat T}\right),$ with components $\tilde p_k = \tilde f_k(o, c)$. 

\subsection{Uncertainty-Aware Action Selection}
\label{sec: uaas}

The greedy action selection approach of equation \eqref{eq: standard policy} does not make use of the calibrated confidence scores from \Cref{sec: calibration}, and overlooks important structure of the action space. Specifically, distinct elements in $\calA$ may represent similar actions. This structure should be leveraged in the manipulation problem, as many tasks can be completed with distinct, but similar actions, e.g. a robotic gripper can open a drawer by pulling at different positions of the handle. 

We propose \Cref{alg: uncertainty aware action selection v2} to remedy the above shortcoming. Rather than selecting the action with the highest confidence score, \Cref{alg: uncertainty aware action selection v2} selects the action for which the neighboring region of potential actions has the highest sum of confidence scores. The neighboring region of an action $a \in \calA$ is defined as
$
\curly{a': d(a, a') \leq \tau},
$
where the metric $d$ is selected based on the problem at hand to capture similarity between two actions in $\calA$, and $\tau$ is a similarity threshold. While both the metric $d$ and the threshold $\tau$ are hand-designed in this work, they could be chosen in a data-driven manner in future work. Note that while the greedy action selection approach in equation \eqref{eq: standard policy} is only sensitive to the ordering of highest confidence score output by the model, \Cref{alg: uncertainty aware action selection v2} needs a well-calibrated model to adequately determine the confidence of the neighboring region.  

\begin{algorithm}
    \caption{Uncertainty-Aware Action Selection}
    \label{alg: uncertainty aware action selection v2}
    \begin{algorithmic}[1]
        \State \textbf{Input: } Calibrated model $\tilde f$, conditioning $c$, observation $o$, robustness threshold $\tau$ 
        \State Compute  $\tilde p_1, \dots, \tilde p_{\abs{\calA}} = \tilde f(o, c).$ 
        
        \State \label{line: action selection}Set $\hat a = \argmax_{a \in \calA} \sum_{a' \in \calA}\tilde p_{j(a')} \mathbf{1}(d(a, a') < \tau)$ 
        
        \State \textbf{Output: }  Uncertainty-aware action $\hat a$
    \end{algorithmic}
\end{algorithm}

\begin{table}

\centering
\begin{tabular}{|c|c|c|c|}
\hline
\textbf{Model} & \textbf{Baseline} & \textbf{Uncalibrated UA} & \textbf{Calibrated UA}\\ \hline
PerAct & $0.382\pm0.012$ & $0.385\pm0.012$ & $\boldsymbol{0.414\pm0.012}$\\ \hline
RVT & $0.602\pm0.012$ & $0.607\pm0.012$ & $\boldsymbol{0.623\pm0.011}$ \\ \hline
CLIPort & $0.803\pm0.005$ & $\boldsymbol{0.833\pm0.005}$ & $\boldsymbol{0.833\pm0.005}$\\ \hline

\end{tabular}
\caption{Summary of results for calibrated and uncalibrated uncertainty-aware approach and the uncertainty oblivious (baseline) approach on  PerAct, RVT, and CLIPort. For PerAct and RVT, the values in the table represent the portion of successfully completed tasks. For CLIPort, the values represent the mean reward, where the maximum reward for an episode is one if the desired task is completed successfully. Standard errors are shown. }
\label{tab: results summary}
\vspace{-16pt}
\end{table}

In Algorithm \ref{alg: uncertainty aware action selection v2}, our method assumes to take a calibrated model $\tilde f$, a language conditioning of the task $c$, the observation $o$, and the robustness threshold $\tau$ as input. At each time step, the calibrated model $\tilde f$ produces a vector of logits $\tilde p$, based on task conditioning and observation. Then, the approach calculates the uncertainty-aware score for the actions by aggregating the neighboring confidence scores of each candidate, where two actions $a$ and $a'$ are considered ``neighbors'' if $d(a, a') < \tau$. Finally, the model makes the decision about the action based on the highest confidence score produced in the previous step.

Note that the approach for selecting the action from the calibrated model in \Cref{line: action selection} of \Cref{alg: uncertainty aware action selection v2} is just one possibility for aggregating the confidence of neighboring actions. An alternative is to convolve the output of the calibrated model with a Gaussian kernel. This approach is discussed in \Cref{s: rvt implementation details}. Another alternative is to construct prediction sets using conformal prediction, and then select a representative from the set. 

A key feature of the proposed approach is that the weights of the pre-trained model are not updated, as it can be both computationally and statistically prohibitive to fine-tune the model.
As a result, the efficacy of an agent using the proposed approach is still bottle-necked by the quality of the pre-trained agent. We therefore expect improvements in the performance from leveraging uncertainty-aware decision making to be limited to some degree by the quality of the pre-trained model. In our experiments, we highlight several cases where the performance boost is most notable, and offer explanations for why this may be the case.

\section{Experimental Results}
\label{sec:result}

We evaluate the method discussed in \Cref{sec: method overview} by deploying it on three pre-trained models: the PerAct model \citep{shridhar2023perceiver}, the RVT model \citep{goyal2023rvt}, and the CLIPort model \citep{shridhar2022cliport}. PerAct and RVT are evaluated on the RLBench dataset \citep{james2020rlbench}, while CLIPort is evaluated in the Ravens benchmark \citep{zeng2021transporter}. Each of these datasets breaks down the desired behavior of the robot into tasks. A task consists of a small collection of related language conditionings. For example, the RLBench task ``stack blocks'' consists of language-conditionings:\footnote{See \cite{james2020rlbench} for all language-conditionings of all tasks.}
\begin{itemize}
    \item ``place three of the blue cubes on top of each other''
    \item ``build a tall tower out of four green cubes''
    \item ``place two of the red cubes on top of each other''
\end{itemize}
Within each task for both the RLBench dataset and the Ravens dataset, the variations of language-conditionings are sampled uniformly at random from the set of task-specific variations built into the environments. We therefore calibrate the models for particular tasks by obtaining expert demonstrations corresponding to language-conditionings sampled uniformly at random from a particular task.

PerAct and RVT were trained on a dataset consisting of 100 expert demonstrations for each of 18 distinct RLBench tasks. We evaluate our approach on each model on all 18 of the tasks they were trained on. For each task, we calibrate the model using temperature scaling with a calibration set of $25$ expert demonstrations, and evaluate on $100$ episodes. 

CLIPort was trained on 1000 demonstrations from each of 9 distinct tasks from the Ravens dataset. We evaluate our approach on four of the nine tasks that the model was trained on (discussed in \Cref{s: UAA benefit}). On each  task, we calibrate the model on 100 expert demonstrations using temperature scaling. Evaluations are performed on 1000 episodes.\footnote{For further details about the implementation of uncertainty-aware action selection on these models, see \Cref{s: model details} and \Cref{s: implementation details}.}

\begin{table}
\centering
\begin{tabular}{|c|c|c|}
\hline
\textbf{Task} & \textbf{Baseline} & \textbf{UA} \\ \hline
\begin{tabular}{@{}c@{}}stack-block-pyramid-seq \\ seen-colors\end{tabular} & $0.9475\pm0.0047$ & $\boldsymbol{0.9815\pm0.0027}$ \\ \hline
\begin{tabular}{@{}c@{}}packing-seen-google \\ objects-seq\end{tabular} & $0.8090\pm0.0104$ & $\boldsymbol{0.8275\pm0.0100}$ \\ \hline
\begin{tabular}{@{}c@{}}packing-seen-google \\ objects-group\end{tabular} & $0.8846\pm0.0080$ & $\boldsymbol{0.9108\pm0.0072}$ \\ \hline
\begin{tabular}{@{}c@{}}assembling-kits-seq \\ seen-colors\end{tabular} & $0.5706\pm0.0116$  & $\boldsymbol{0.6134\pm0.0115}$ \\ \hline
\textbf{Task with Distractors} & \textbf{Baseline} & \textbf{UA} \\ \hline
\begin{tabular}{@{}c@{}}stack-block-pyramid-seq \\ seen-colors-diff-sizes\end{tabular} & $0.7963\pm0.0073$ & $\boldsymbol{0.9711\pm0.0031}$ \\ \hline
\begin{tabular}{@{}c@{}}assembling-kits-seq \\ seen-colors-diff-sizes\end{tabular} & $0.4832\pm0.0028$ & $\boldsymbol{0.5544\pm0.0030}$ \\ \hline
\end{tabular}
\caption{Task-specific mean rewards for CLIPort on four tasks from the original Ravens dataset and two tasks with small distractors added to the original tasks.}
\vspace{-12pt}
\label{tab: cliport success Results}
\end{table}

\begin{figure*}[t]
      \includegraphics[height=9cm]{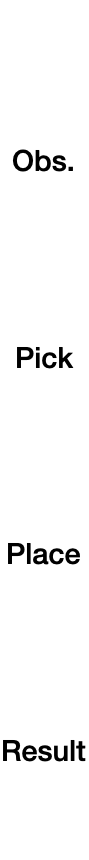}
    \subfloat[Put Gray Block on the 
Lightest Brown Block]{\label{fig:stack_blocks}
      \includegraphics[height=9cm]{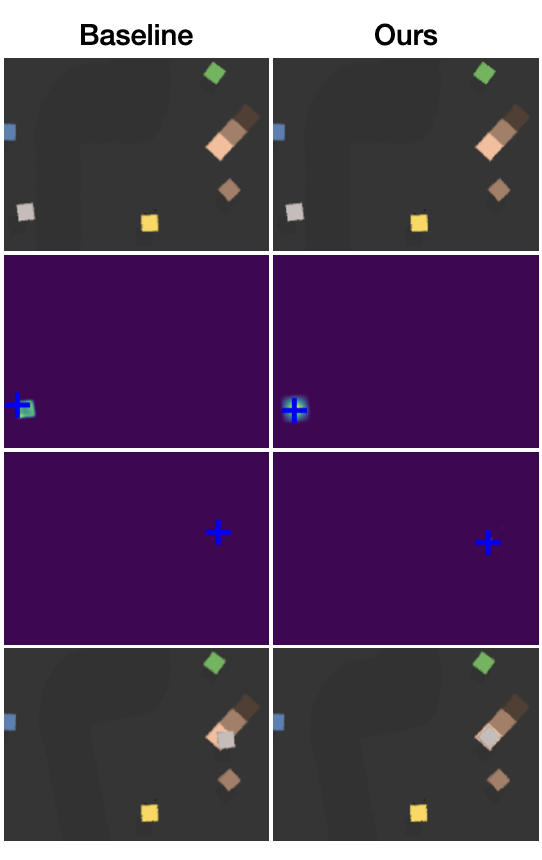}}
    \hspace{0.6em} 
    \subfloat[Pack All the Spatula with Purple 
Head Objects in the Brown Box]{\label{fig:cliport_kit_assembly1}
      \parbox[b]{0.3\linewidth}{\centering \includegraphics[height=9cm]{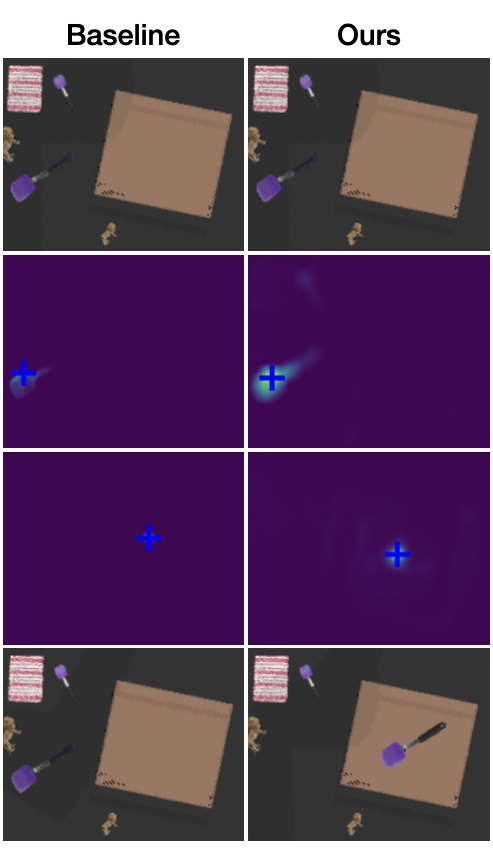}\\
      }} 
    \hspace{0.1em} 
    \subfloat[Put the Brown Ring
In the Brown Box]{\label{fig:cliport_kit_assembly2}
      \parbox[b]{0.3\linewidth}{\centering \includegraphics[height=9cm]{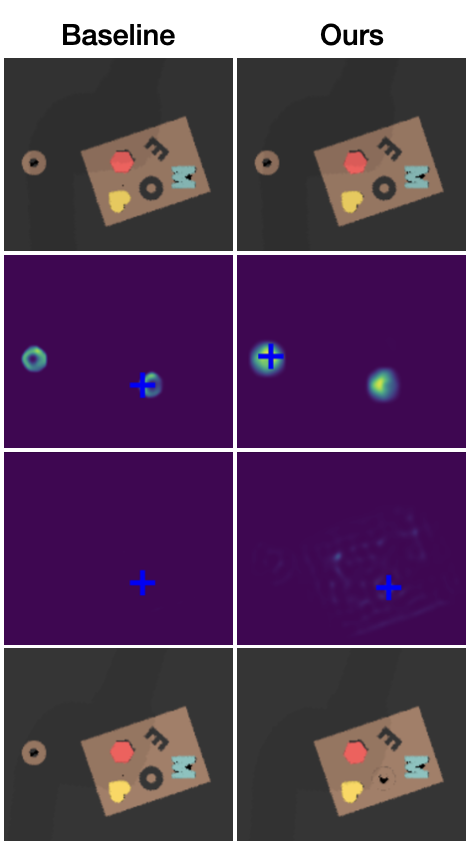}\\
      }} 
    \caption{Qualitative results for CLIPort with and without uncertainty-aware action selection. The language commands are in the subfigure captions. From top to bottom, each row represents the model's observation, the heatmap in pick phase, the heatmap in place phase, and the observation after executing the actions. In \ref{fig:stack_blocks} and \ref{fig:cliport_kit_assembly1}, our method forces the model to choose a central location on the target object instead of the edges. In \ref{fig:cliport_kit_assembly2}, our method corrects the model by smoothing out the action with highest raw score.}
    \label{fig: cliport examples}
    \vspace{-12pt}
\end{figure*}

The results from running uncertainty-aware action selection on the three models are summarized in \Cref{tab: results summary}. We see that the uncertainty-aware approach offers a benefit to the overall task completion rate for all models. To investigate exactly when the approach provides a benefit, the subsequent sections study three pertinent questions, summarized below.
\begin{enumerate}
    \item \textbf{\Cref{s: UAA benefit}: Why does uncertainty-aware action selection help?} Uncertainty-aware action selection prevents the end effector from moving to positions described by isolated, high-confidence locations output by the pre-trained model. This can prevent the gripper from completely missing the object, and encourages a better grip when it is successful in gripping the object. 
    \item \textbf{\Cref{s: calibration section}: Is calibration necessary?} It depends on the model. For simple models trained on very large datasets (CLIPort), the pre-trained model appears to be fairly well calibrated, and calibration is not necessary to achieve the benefit of uncertainty-aware decisions. Both PerAct and RVT are poorly calibrated on any given task, and calibrating the model is critical to achieve the benefit of uncertainy-aware decisions. 
    \item \textbf{\Cref{s: distractors}: Can uncertainty-awareness enable generalzation to new settings?} 
    We demonstrate that small distractor objects to the scene can have a devastating performance of the pre-trained model. However, uncertainty-aware action selection achieves essentially the same performance it does in the absence of distractors. This provides a strong indication that accounting for uncertainty-aware action selection can improve generalization to certain task distribution shifts.
\end{enumerate}

\subsection{Why Does Uncertainty-Aware Action Selection Help?}
\label{s: UAA benefit}

\begin{figure*}
    \centering    \includegraphics[width=0.8\linewidth]{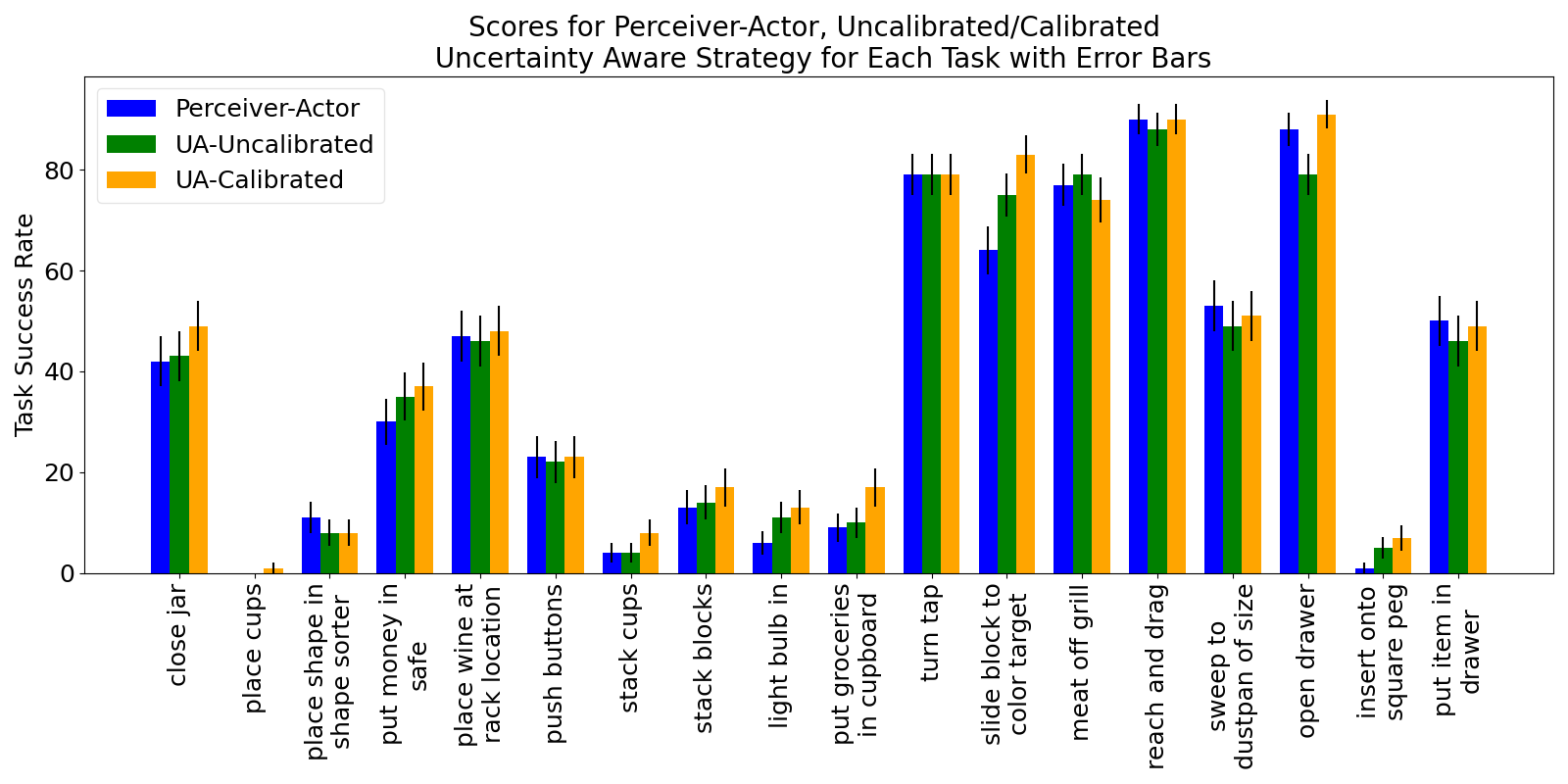}
    \caption{Success rate score for Perceiver-Actor, uncertainty-aware strategy with calibrated/uncalibrated confidence for each task}
    \label{fig:PerAct per task}
    \vspace{-12pt}
\end{figure*}

\begin{figure*}
    \centering
    \includegraphics[width=0.8\linewidth]{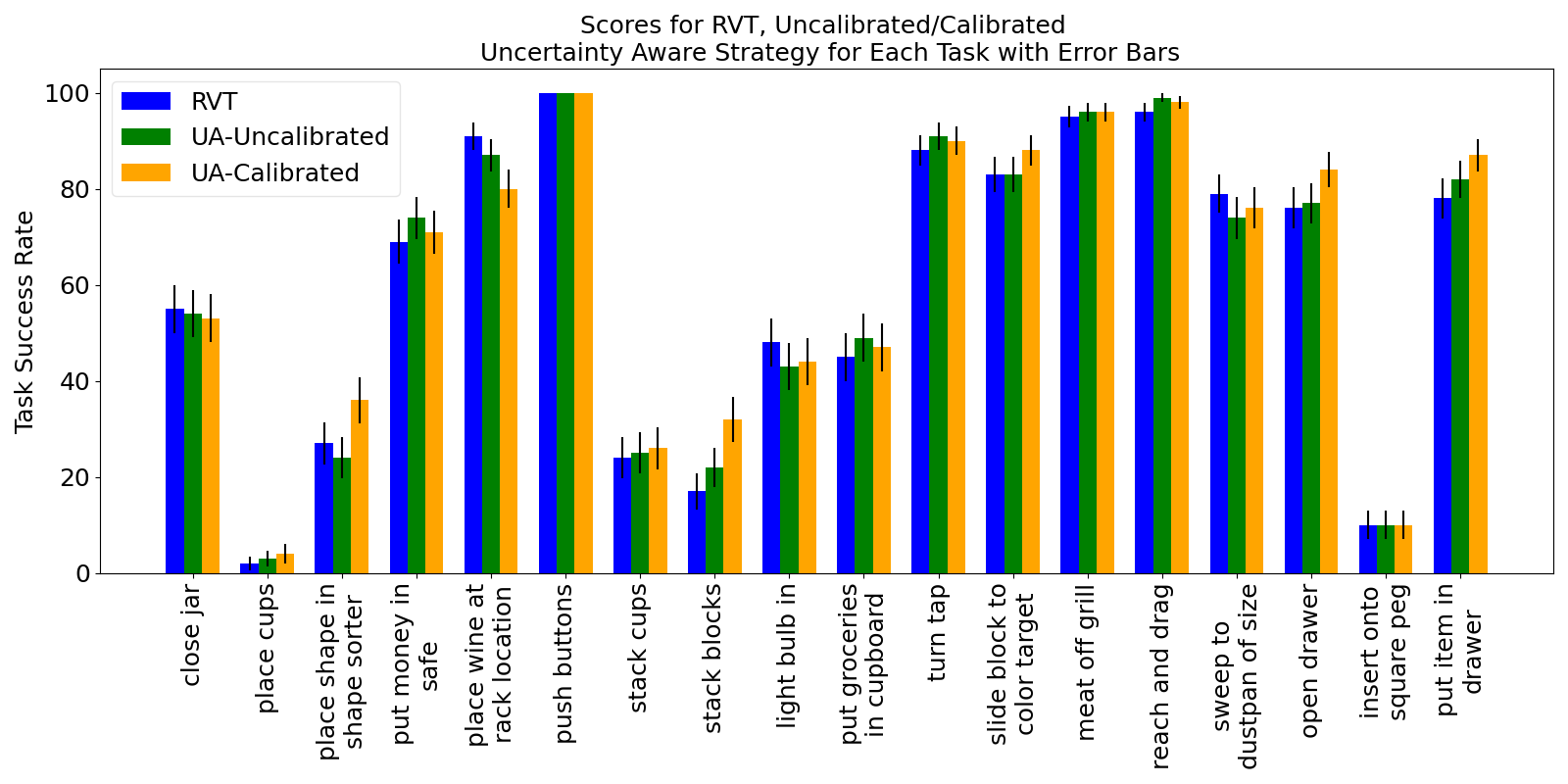}
    \caption{Success rate score for RVT, uncertainty-aware strategy with calibrated/uncalibrated confidence for each task}
    \label{fig:RVT per task}
    \vspace{-12pt}
\end{figure*}

To investigate why uncertainty-aware action selection helps, we first look at the per-task success rates of the CLIPort model. In the first four rows of \Cref{tab: cliport success Results}, we compare uncertainty-aware action selection with the baseline on four tasks from the Ravens dataset. 
From the results, we find that incorporating uncertainty is particularly useful when the CLIPort task requires precise placement of target objects. This is important for stacking blocks, packing objects, and assembling kits. 
The reason that uncertainty awareness is beneficial in these settings is that it enables conservative decisions in the picking phase by encouraging the end effector to pick up the object at the center instead of at the edges, as illustrated in \Cref{fig: title figure}, \Cref{fig:stack_blocks}, and \Cref{fig:cliport_kit_assembly1}. 

In these examples, our method takes the calibrated score map of the baseline (left column of the second row), and generates uncertainty-aware score calculation (right column of the second row). Conditioned on the more conservative picking location, the agent picks up the object more easily and finds a better placing location, as shown in the third row. The bottom row of each step shows the result of the manipulation at the current time step.

Additionally, by smoothing confidence scores across possible actions, the uncertainty-aware approach prevents the end effector from selecting isolated high-confidence actions. The benefit of this is illustrated in the second row of \Cref{fig:cliport_kit_assembly2}.

\begin{figure}[t]
    \centering
    \includegraphics[width=0.9\linewidth]{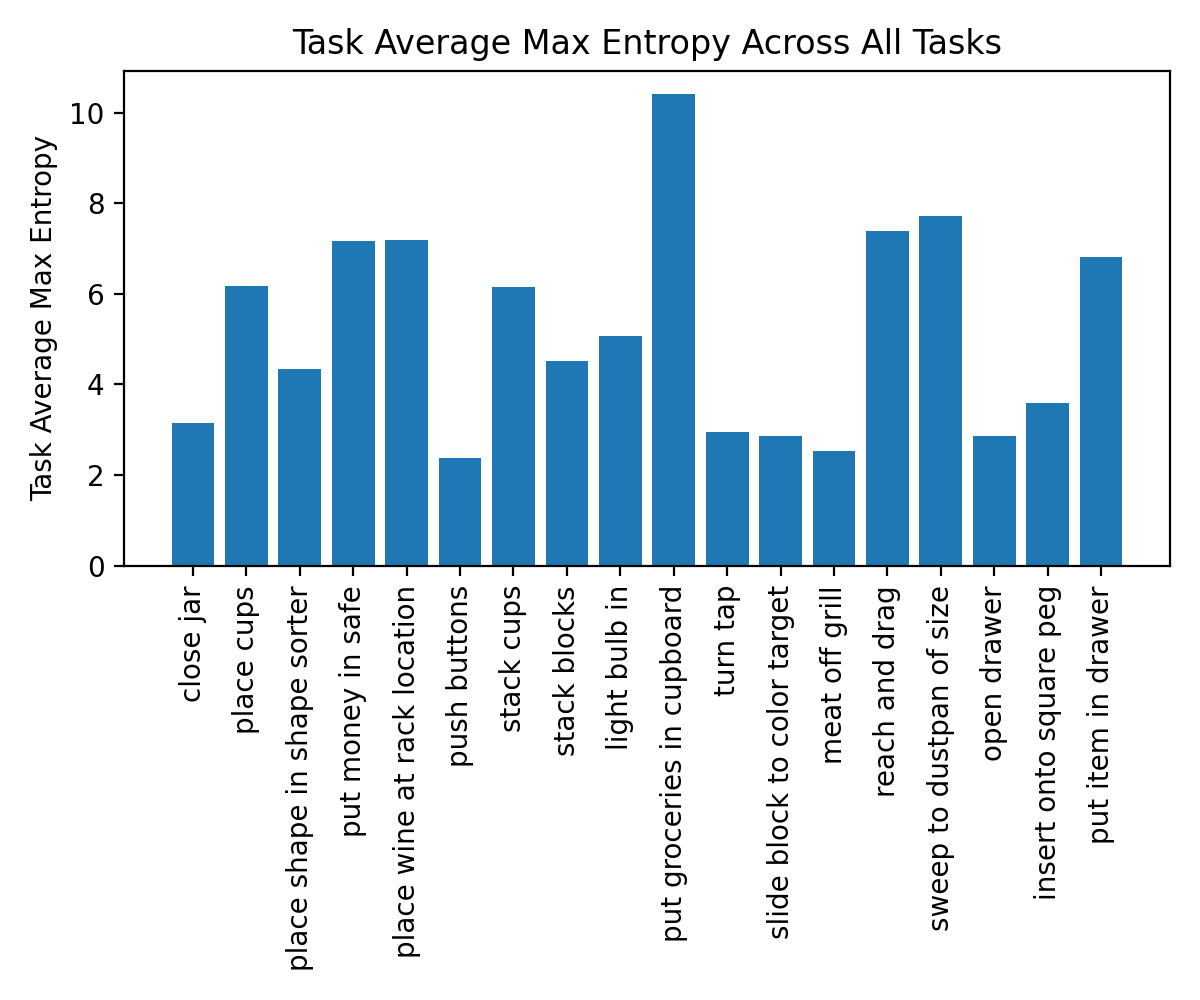}
    \caption{Maximum entropy over all episodes in each task of PerAct}
    \label{fig: max entropy peract}
    \vspace{-16pt}
\end{figure}

The per task success rates of PerAct are shown in \Cref{fig:PerAct per task} while the per task success rates of RVT are shown in \Cref{fig:RVT per task}. Due to the higher complexity of the tasks from RLBench and the higher standard error in the results, it is difficult to draw concrete conclusions about when the benefit of accounting for uncertainty is most important. However, we do see the most pronounced improvement on tasks with a lower success rate. This benefit appears to be due to the larger discrepancy between the action with the highest confidence score and the action with the largest sum of neighboring confidence scores on tasks for which the model is overall less confident. In particular, we plot the maximum entropy of the translation distribution output by PerAct for each task in \Cref{fig: max entropy peract}. We see that for the push buttons task, it has the highest success rate and the lowest maximum entropy. Meanwhile, the highest maximum entropy comes from the task put groceries in cupboard, upon which we see a significant boost coming from our method.

In light of the above experimental results, we conclude that by encouraging the model to select actions with higher neighboring confidence scores, our method prevents the model from choosing aggresive action candidates, thereby resulting in better grasping performance.

\subsection{Do we need calibration?}
\label{s: calibration section}

We see in \Cref{tab: results summary} that for both PerAct and RVT, the uncertainty-aware action provides only a very small improvement in the absence of calibration (0.3\% and 0.5\% boost in task success rate, respectively); however, they yield a significant task success rate if a calibration step is included (3.2\% and 2.1\% boosts, respectively).

\begin{figure}
    \centering
    \begin{minipage}{\linewidth}
    \subfloat[uncalibrated]{\label{fig: uncalibrated}
    \includegraphics[width = 0.9\linewidth]{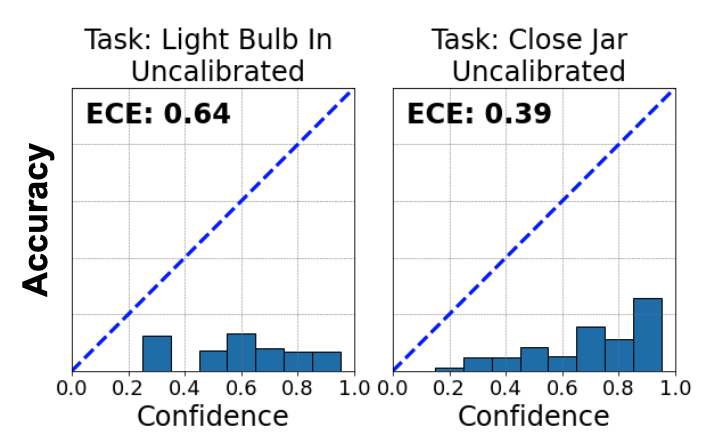}
    }
    \end{minipage}
    \begin{minipage}{\linewidth}\vspace{0.1in}
    \subfloat[Calibrated]{\label{fig: calibrated}
    \includegraphics[width = 0.9\textwidth]{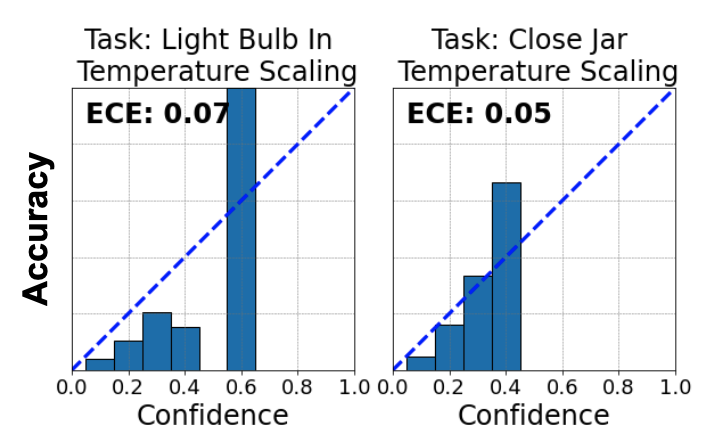}
    }
    \end{minipage}
      
    \caption{Reliability diagrams are shown for PerAct on two of the RLBench tasks. \Cref{fig: uncalibrated} illustrates the reliability diagrams before calibration, while \Cref{fig: calibrated} shows the reliability diagrams after running temperature scaling on a calibration set of 25 expert demonstrations from the each task. Overlayed on the plot is the value for the expected calibration error (ECE, see \citet{guo2017calibration} for further details). We see that temperature scaling provides a marked improvement in the calibration. } 
    \vspace{-16pt}
    \label{fig: reliability diagram}
\end{figure}

To understand why this is the case, we construct reliability diagrams in \Cref{fig: reliability diagram}. The figure depicts this reliability diagram for PerAct on two tasks, with both the uncailbrated model and the model calibrated using temperature scaling. The $y$-axis is the model accuracy (the correctness likelihood) at each confidence level, which is the number of predicted actions that are the same as the expert demonstration, divided by the total number of predictions made with confidence within the thresholding interval. The $x$-axis depicts the model confidence, which is maximum value in the probabilty distribution representing the calibrated model:  $\max_{a\in\calA} \tilde f_a(o,c)$. The confidence is perfectly calibrated when the accuracy and the confidence are linearly related with slope of one. For more details on reliability diagrams, see \citet{guo2017calibration}. These diagrams illustrate the value of temperature scaling for improving model calibration using a calibration set of 25 expert demonstrations from the task of interest. From these results, we conclude that PerAct has very poorly calibrated uncertainty measurements, but applying temperature scaling using a small number of demonstrations from a specific task improves calibration substantially. 

The importance of this calibration is further reflected by the improvement offered by the calibrated uncertainty-aware approach over the uncalibrated uncertainty-aware approach in \Cref{tab: results summary}. As in \Cref{tab: results summary}, RVT benefits from temperature scaling. However, CLIPort does not see a benefit of temperature scaling in improving the success rate of uncertainty-aware action selection. We find that the calibrating cross-entropy loss barely changes during from calibration, indicating that the pre-trained model is already fairly well-calibrated for each of the individual tasks.

\subsection{Can Uncertainty-Awareness Improve Generalization?}
\label{s: distractors}


To determine whether uncertainty-aware decisions can help pre-trained models generalize to out-of-distribution settings, we designed two new tasks. One of these is a modification of stack blocks, and the other one is a modification of the assemble kits task. Both of the new tasks introduce a collection of smaller-sized target objects of various colors and shapes, which serve as distractors to the model. In the last two rows of \Cref{tab: cliport success Results},  we find that the pre-trained method is very fragile to such distractors, and they cause a massive drop in performance from 94.75\% completion to 79.63\% for stacking blocks, and from 57.06\% to 48.32\% for assembling kits. However, the uncertainty-aware approach is robust to such distractors: performance degrades only from 98.15\% and 61.34\% completion to 97.11\% and 55.44\%, respectively. 

\begin{figure}[t]
    \centering
    \includegraphics[width=0.7\linewidth]{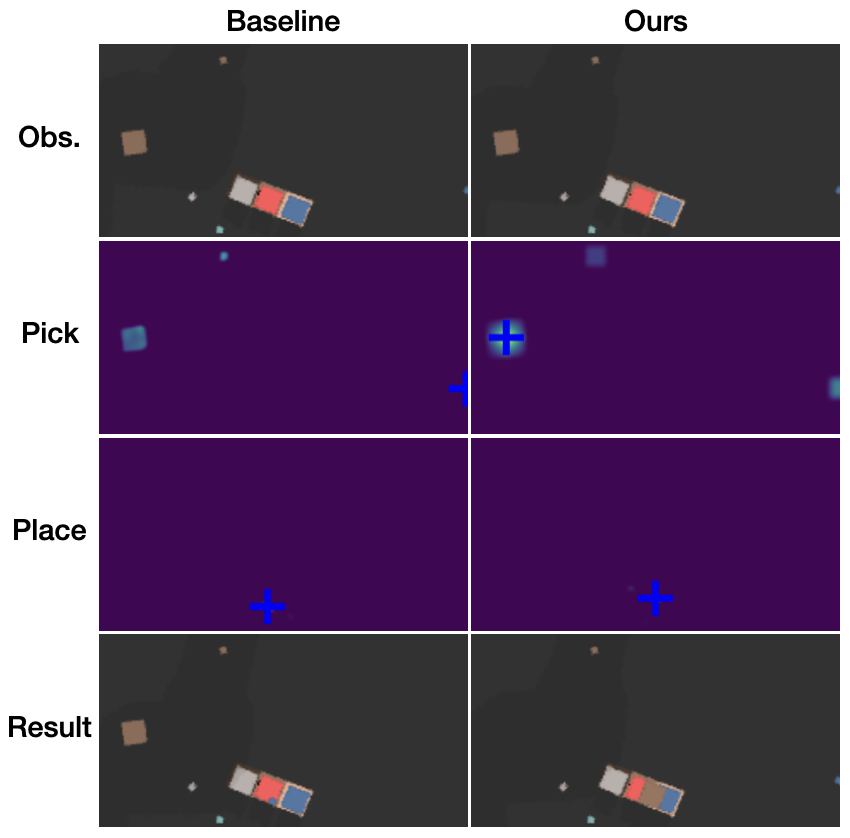}
    \caption{Qualitative results of CLIPort on instruction "Put the brown block on the blue and red blocks" when distractors are introduced.}
    \vspace{-16pt}
    \label{fig: distractor example}
\end{figure}

In \Cref{fig: distractor example}, we demonstrate the impact of the distractors on the pre-trained model. We see that the precense of distractors causes the pre-trained model to pick up one of the distractor blocks which has a single pixel of high confidence. The uncertainty-aware approach smooths the heatmap, thereby determining that the isolated high confidence event is due to a distractor, and that the best course of action is to select a region where the model output a collection of neighboring actions with high confidence.  

More generally, we believe these results indicate the potential value of incorporating uncertainty estimates for decision making in out-of-distribution scenarios.

\section{Conclusion}
\label{sec:conclusion}

In this paper, we proposed the use of simple classification 
model calibration techniques in conjunction with an uncertainty-aware action selection approach to improve the success rates for language-conditioned imitation learning in robotic manipulation. Our results suggest that by appropriately calibrating the model and incorporating an uncertainty-aware decision making strategy, we can benefit the pre-trained ``foundation models'' in both seen and unseen tasks. This approach does not require any additional training or fine-tuning of the original model and is scalable to versatile robot manipulation models and platforms.

The modest, though notable performance boost that occurs by incorporating uncertainty-aware decisions for pre-trained models highlights the potential for such approaches for generalist robotic policies in general. An exciting avenue for future work is to obtain calibrated uncertainty models for generalist dynamics forecasting which can then be used as part of an uncertainty-aware trajectory planning or model predictive control scheme. 



\appendix

\subsection{Model details}
\label{s: model details}

For PerAct and RVT, the action space consists of three coordinates for the discretized desired end effector translation, three coordinates for the discretized desired end effector rotation, and two binary values indicating gripper status and collision avoidance module activation. The desired action is then achieved using a sampling-based motion planner. The world state is represented by a voxel observation of the environment. PerAct directly takes this voxel obseration as an input (see \cite{shridhar2023perceiver}), while RVT takes five RGB-D virtual camera projections of this voxel observations as input (see \citet{goyal2023rvt}). Due to differences between the output of each model, and the structure  assumed by our approach in \eqref{eq: standard policy} and due to computational challenges of \Cref{alg: uncertainty aware action selection v2} in high dimensions, we make several model-specific adjustments to the uncertainty-aware action selection. See \Cref{s: peract implementation details} and \Cref{s: rvt implementation details} for PerAct and RVT, respectively. 

CLIPort instead breaks tasks into a sequence of pick-and-place actions. Each pick action is described by the two dimensional end effector location, while each place action is described by the two dimensional end effector location and the one dimensional end effector orientation. For each pick and place step, the CLIPort model takes as input the language description of the task, and a top-down virtual RGB-D image of the scene. See \citet{shridhar2022cliport, zeng2021transporter} for further details.  Note that both picking and placing outputs a two dimensional heatmap of confidence scores for the appropriate end effector location. Applying \cref{alg: uncertainty aware action selection v2}, we calculate the uncertainty-aware heatmap and determine the appropriate picking location. For the orientation during the place step, we directly trust the output of pre-trained model. 

All hyperparameters are determined by performing a parameter sweep using cross validation. 

\vspace{-12pt}

\subsection{Additional Implementation Details}
\label{s: implementation details}
\subsubsection{PerAct} \label{s: peract implementation details}
The action space $\calA$ is very large, and it is computationally expensive to search over the entire action space to calculate the neighboring confidence for each action. Meanwhile, while the probability output from a well-trained model can span the entire action space, there are relatively few entries with high confidence scores. For these reasons, we restrict our uncertainty-aware action selection search to translation actions that have relatively high confidence scores. For rotation, gripper status, and collision avoidance module activation, we directly trust the pre-trained model.

To restrict the action search space, we make several modifications to \Cref{alg: uncertainty aware action selection v2}. These are summarized in \Cref{alg: uncertainty aware action selection detailed}. Our method assumes that we have a calibrated model $\tilde f$. First we apply a probability threshold to filter out the actions with probabilities higher than this threshold. Then, if the resulting action space has more than k elements, we keep only the top k of them. After this, we select the center of the remaining actions, define a search space $\calH$ around the center with radius $D/2$, and go loop through the actions in the search space. For each action $a$, we extract the remaining actions $a' \in \calA'$ around it with $d(a,a') < \tau$, and calculate the accumulative scores for $a$. Finally, we take the action with the greatest accumulative score to be our desired action. This mitigates the computational problem described above by overlooking non-informative parts in the model's output. We choose $\alpha=1/\abs{\calA}$ and $k=4000$.

\begin{algorithm}
    \caption{Detailed UA Action Selection Strategy}
    \label{alg: uncertainty aware action selection detailed}
    \begin{algorithmic}[1]
        \State \textbf{Input: } Calibrated model $\tilde f$, conditioning $c$, observation $o$, threshold $\tau$, probability threshold $\alpha$, search size $D$.
        \State Compute action probabilities: $\tilde p_1, \dots, \tilde p_{\abs{\calA}} = \tilde f(o; c).$
        \State $\calA' = \{a \in \calA, \ \tilde p_a > \alpha \}$, 
        \State if $\abs {\calA'} > k$: keep $k$ actions in $\calA'$ with the highest scores
        \State c = mean($a' \in \calA'$)
        \State \calH = \{ $a \in [c-D/2, c+D/2]^3 \cap \calA$ \}
        \State Set $\hat a = \argmax_{a \in \calH} \sum_{a' \in \calA'} \tilde p_{j(a')} \mathbf{1}(d(a, a') < \tau)$
        \State \textbf{Output: }  Uncertainty-aware action $\hat a$
    \end{algorithmic}
\end{algorithm}

\subsubsection{RVT} 
\label{s: rvt implementation details}
RVT is modified from PerAct to predict the heatmap for the translation in virtual image planes instead of the raw 3D voxel space. As such, we adjust \Cref{alg: uncertainty aware action selection v2} to work with the 2D output accordingly. For each heatmap, we convolve it with a Gaussian blurring kernel to get an uncertainty-aware heatmap and reconstruct a 3D uncertainty-aware point cloud.\footnote{A Gaussian blurring kernel is a generalization of the weighted sum in \Cref{line: action selection} from \Cref{alg: uncertainty aware action selection v2}. The width and standard deviation of this kernel are treated as hyperparameters that are determined via cross validation. We apply it only in this setting because the convolution becomes only a 2D convolution, which is necessary in the image space.}  We apply this convolution strategy in 1D space for each axis of rotation. Consequently, the uncertainty-aware action decision is made over the resulting uncertainty-aware heatmaps for both translation and rotation.




\bibliographystyle{IEEEtranN}
\bibliography{refs}

\end{document}